\pdfoutput=1

\documentclass[11pt]{article}

\usepackage[]{EMNLP2022}

\usepackage{times}
\usepackage{latexsym}

\usepackage[T1]{fontenc}

\usepackage[utf8]{inputenc}

\usepackage{microtype}
\usepackage{microtype}
\usepackage{graphicx}
\usepackage{amsmath}
\usepackage{booktabs}
\usepackage{url} 
\usepackage{xcolor}
\usepackage{latexsym}
\usepackage{amsfonts}
\usepackage{amsmath}
\usepackage{bbm} 
\usepackage{amssymb}
\usepackage{subfigure}
\usepackage{balance}
\usepackage{threeparttable}
\usepackage{multirow}
\usepackage{float}
\usepackage{longtable}
\usepackage{lipsum}
\usepackage{inconsolata}

\title{\emph{Step out of KG}: Knowledge Graph Completion via \\ Knowledgeable Retrieval and Reading Comprehension}

  \newcommand*{\email}[1]{\texttt{#1}}

  \author{
    \textbf{Xin Lv}$^{1,2}$, \textbf{Yankai Lin}$^{3}$, \textbf{Zijun Yao}$^{1,2}$, \textbf{Kaisheng Zeng}$^{1,2}$ \\ \textbf{Jiajie Zhang}$^{1,2}$,  \textbf{Lei Hou}$^{1,2}$, \textbf{Juanzi Li}$^{1,2}$\\
    $^1$Department of Computer Science and Technology, BNRist \\
    $^2$KIRC, Institute for Artificial Intelligence, Tsinghua University, Beijing 100084, China \\
    $^3$Renmin University of China, Beijing 100872, China \\
    \email{lv-x18@mails.tsinghua.edu.cn}
    }

\begin{document}
\maketitle
\begin{abstract}
Knowledge graphs, as the cornerstone of many AI applications, usually face serious incompleteness problems. In recent years, there have been many efforts to study automatic knowledge graph completion (KGC), most of which use existing knowledge to infer new knowledge. However, in our experiments, we find that not all relations can be obtained by inference, which constrains the performance of existing models. To alleviate this problem, we propose a new model based on information retrieval and reading comprehension, namely IR4KGC. Specifically, we pre-train a knowledge-based information retrieval module that can retrieve documents related to the triples to be completed. Then, the retrieved documents are handed over to the reading comprehension module to generate the predicted answers. In experiments, we find that our model can well solve relations that cannot be inferred from existing knowledge, and achieve good results on KGC datasets.
\end{abstract}

\section{Introduction}
\label{sec:introduction}

Knowledge Graphs (KGs), which represent knowledge as structured triples, are the infrastructure for many AI studies.
However, most real KGs face serious incompleteness problems. For example, about 71\% of people in Freebase~\cite{Freebase} lack birthplace information~\cite{KGVault}, which limits the performance of downstream tasks.

To alleviate the incompleteness of KGs, knowledge graph completion (KGC) task is proposed, which usually uses the schema of KGs to determine which knowledge is missing and use KGC models to complete it. Among these models, knowledge graph embedding (KGE) models~\cite{TransE} are dominant, which usually embed entities and relations into vector spaces and predict missing knowledge based on vector operations.

\begin{table}[t]
    \centering
    \small
    \scalebox{0.98}{
    \begin{tabular}{lcc}
    \toprule[1pt]
    \textbf{Relation} &  MRR & Hits@10 \\
    \midrule[0.5pt]
    {\color{red} \textbf{\textit{cause of death}}}  & .099 & .162\\
    {\color{red} \textbf{\textit{/people/person/place\_of\_birth}}}  & .075 & .134\\
    \midrule[0.5pt]
    \textit{ethnic group}  & .452 & .684\\
    \textit{/people/ethnicity/languages\_spoken}  & .325 & .738\\
    \bottomrule[1pt]
    \end{tabular}
    }
    \caption{\label{tab:different_relations} Performance gap of KGE model TuckER~\cite{TuckER} on different relations. The bolded red relations are difficult to infer from existing knowledge and the KGE model struggles with its performance.
    } 
  \end{table}

However, the effectiveness of KGE models relies on the assumption that the missing knowledge in the KGs can be inferred from existing knowledge. 
Although this assumption holds for most relations in the KGs, there still exist several non-negligible exceptions.
We refer to these exception relations as uninferable relations and the others as inferable relations. For example, for the relation \textit{cause of death}, it is difficult to infer the cause of someone's death from what we already know. 
Table~\ref{tab:different_relations} gives the performance of the KGE model on different relations, which shows that the KGE model struggles with the uninferable relations.

A reasonable solution to complete the uninferable relations is to extract the knowledge from the corresponding text. There are two main types of related KGC models: (1) KGE models that introduce description information of entities~\cite{StAR}; (2) models based on pre-trained language model (PLM)~\cite{KGT5}. However, both types of models have corresponding drawbacks, where the former cannot guarantee that the knowledge to be completed is in the description and the latter depend on the knowledge contained in the PLM. In addition to KGC models, relation extraction (RE) models~\cite{levy2017zero} can also derive knowledge from text. But they do not apply to this task since they require the corresponding text, which is missing in the KGC task.

To complete uninferable relations more accurately, we are inspired by Open-domain Question Answering (OpenQA) models~\cite{REALM,RAG} and propose a novel KGC method based on information retrieval and machine reading comprehension, namely IR4KGC.
Specifically, the triple query $(h, r, ?)$ is firstly converted to a search query containing its knowledge semantics. Then, our model uses a pre-trained knowledge retrieval module to retrieve documents that match the search query and generates the final predictions based on a generative PLM. Besides, the retrieved documents provide additional interpretability.

Most of the existing OpenQA models are based on Dense Retrieval~\cite{DPR} or BM25~\cite{BM25}. These retrieval modules can retrieve natural language queries well, but it is challenging to handle search queries containing rich knowledge semantics in the KGC task. 
To solve this problem, we construct a training corpus for retrieval based on the idea of distant supervision and pre-train our knowledge retrieval module on the task KGC. Thus it can better capture the knowledge semantic information contained in the search query and return more relevant documents.

Experimental results on two KGC datasets show that IR4KGC achieves superior results on uninferable relations over the KGE models. In addition, the combination of IR4KGC and the KGE model achieves the best performance on all datasets.

\section{Related Work}

\subsection{Knowledge Graph Completion}

KGE models are the main components of knowledge graph completion models. KGE models can be divided into four main categories: (1) translation-based models~\cite{TransE,RotatE};
(2) models based on tensor decomposition~\cite{RESCAL,TuckER}; (3) models based on neural networks~\cite{NTN,ConvE}; (4) models that introduce additional information~\cite{KR-EAR,StAR}. As we introduced in Section~\ref{sec:introduction}, KGE models struggle with uninferable relations.

There are some PLM-based KGC models being proposed in recent years, most of which use PLM to determine the correctness of a given triple~\cite{KG-BERT,PKGC} or to directly generate the predicted tail entities~\cite{KGT5}. 
Implicit knowledge in PLM can help the model to complete uninferable relations. But these models still have drawbacks since it is difficult for PLM to accurately remember all knowledge in the world.

RE models can also complete knowledge from text. However, RE aims to extract all the knowledge from text, and it is difficult to do specific knowledge completion. Furthermore, the text required for RE is also missing, making RE unsuitable for the KGC task in this paper.

\subsection{Open-domain Question Answering}

Open-domain Question Answering aims to answer open-domain questions without context. Most of the OpenQA models in recent years have adopted the retrieving and reading pipeline~\cite{chen2017reading,REALM,RAG}.  Specifically, these models use retrieval modules such as Dense Retrieval~\cite{DPR} or BM25~\cite{BM25} to retrieve relevant documents and give answers using extraction or generation-based methods. However, these retrieval modules are difficult to adapt to KGC tasks and have low retrieval efficiency. In addition, there are some OpenQA models based on knowledge-guided retrieval~\cite{min2019knowledge,asai2019learning}, but they are limited by KGs with Wikipedia links and are difficult to adapt to most KGs.

\section{Method}

Given a triple query $(h, r, ?)$, where $h$ is the head entity and $r$ is the relation, we transform it into a search query and retrieve relevant documents using our retrieval module. After that, the conditional generation module generates predicted answers based on the documents. 
These two modules are optimized jointly following~\citet{RAG}.

\subsection{Knowledge-based Information Retrieval}

\noindent{\textbf{Triple Query Transformation}.} For a triple query $tq = (h, r, ?)$, we have two functions to convert it into a search query, denoted as $\text{F}_L$ and $\text{F}_{LA}$. $\text{F}_L(tq) = \text{LABEL(}h\text{)} \Vert \text{LABEL(}r\text{)}$, where $\text{LABEL(}x\text{)}$ is the label corresponding to $x$ and $\Vert$ denotes the concatenation operation. 
$\text{F}_{LA}$ uses alias to increase the query diversity.
Specifically, $\text{F}_{LA}(tq) = \text{TEXT(}h\text{)} \Vert \text{TEXT(}r\text{)}$, where $\text{TEXT(}x\text{)}$ has a $50\%$ probability of being a label of $x$ and a $50\%$ probability of being a random alias of $x$.

\noindent{\textbf{Pre-training Method}.} Following the training approach of DPR~\cite{DPR}, we pre-train our information retrieval model using the contrastive learning approach. Specifically, we use two separate BERT~\cite{BERT} models as the search query encoder and the document encoder, respectively. We use the dot product to measure the similarity between search query and document. Formally, the similarity can be defined as
\begin{equation}
    sim(sq, d) = \text{QEnc}(sq)^{\top} \cdot \text{DEnc}(d) , 
\end{equation}
where $sq$ and $d$ are search query and document, QEnc and DEnc are the search query encoder and the document encoder, respectively.

We train our encoder using the contrastive learning method. Specifically, for a search query $sq$, we have one positive document and $n$ negative documents, where $n$ is the training batch size. One of the negative documents is the strong negative corresponding to the search query, and the remaining $n-1$ are  negative documents using in-batch strategy~\cite{henderson2017efficient}, i.e., using the positive documents corresponding to the rest of the search query in the same batch as negative documents. The final training loss can be defined as:
\begin{equation}
\small
    L = \sum_{sq \in \mathcal{Q}} -\text{log} \frac{e^{sim(sq, d^+)}}{e^{sim(sq, d^+)} + \sum_{i = 1}^{n}{e^{sim(sq, d^{-}_{i})}}} ,
\end{equation}
where $\mathcal{Q}$ is the search query set for pre-training, $d^+$ is positive document for search query $sq$, $d^{-}_{1}, \dots, d^{-}_{n}$ are negative documents for $sq$.

\noindent{\textbf{Pre-training Corpus Construction}.} Our pre-training corpus for knowledge-based information retrieval is constructed based on Wikidata and Wikipedia. We first construct the search query set $\mathcal{Q}$. After that, for each triple query in $\mathcal{Q}$, we construct the corresponding positive document and strong negative document.

For search query set construction, we first select the triples from Wikidata whose head and tail entities have English Wikipedia links to form the triple set $\mathcal{T}_f$. We then randomly select 600,000 triples from $\mathcal{T}_f$ to form our triple set $\mathcal{T}$ for pre-training. For every triple $(h, r, t)$ in $\mathcal{T}$, we have two triple query $(h, r, ?)$ and $(t, inv\_r, ?)$. We call the $\text{F}_{LA}$ function $N$ times to get their $N$ search queries separately. Finally, all search queries form the set $\mathcal{Q}$.

For each article in Wikipedia, we cut it into several documents of length up to $100$. we end up with $25,570,310$ documents. We also keep the mention information in each document. For every search query $sq$ in $\mathcal{Q}$, if its corresponding triple query is $(h, r, ?)$ and the correct tail entity is $t$, we have three types of positive documents, i.e., entity-type, distant-type and answer-type positive documents. 

The positive documents of \textbf{entity-type} help the retrieval model to better represent the entity information. They can be defined as $\{\text{Adocs}(h) \cap \text{Mdocs}(t)\}$, where $\text{Adocs}(x)$ represents the document set cut out of the article corresponding to $x$ and $\text{Mdocs}(x)$ denotes the set of all documents where $x$ appears as a mention. The positive documents of \textbf{distant-type} help the retrieval model to better represent the relation information by the idea of distant supervision in relation extraction. They can be defined as $\{\text{Mdocs}(h) \cap \text{Mdocs}(t)\}$. The positive documents of \textbf{answer-type} aim to increase diversity and they can be defined as $\{\text{Mdocs}(t)\}$. 
For each search query $sq$ in $\mathcal{Q}$, we first select positive documents of the above three types in the proportion of 45\%, 45\% and 10\%, and then randomly select one of them as the final positive document.

For each search query $sq$, we use the first document without $t$ as mention according to the search ordering of BM25 as the strong negative document.

\noindent{\textbf{Fine-tuning on KGC datasets}.} Before Fine-tuning, we use the document encoder to get embeddings for all 25,570,310 documents and construct an index based on Faiss~\cite{faiss}. Following~\citet{RAG}, during fine-tuning, we only optimize the parameters of the search query encoder. In addition, to increase certainty, we use the $\text{F}_L$ function to transform the triple query.

\begin{table*}[t]
  \centering
  \hspace*{-0.32cm}
\setlength\tabcolsep{5.5pt}
  \scalebox{0.788}{
  \begin{tabular}{lcccccccc}
    \toprule
    \multirow{2}{*} {Model}     & \multicolumn{4}{c}{\textbf{CoDEx-M}} & \multicolumn{4}{c}{\textbf{FB15K-237-N}} \\ 
  \cmidrule(r){2-5}\cmidrule(lr){6-9}
    & MRR & Hits@1 & Hits@3 & Hits@10 & MRR & Hits@1 & Hits@3 & Hits@10  \\ 
  \midrule
  RotatE~\cite{RotatE} & 29.8 & 20.7 & 33.9 & 44.7 & 27.9 & 17.7 & 32.0 & 48.1 \\
  TuckER~\cite{TuckER} & 31.0 & 23.7 & 34.1 & 44.6 & 31.2 & 22.8 & 34.6 & 48.6 \\
  StAR~\cite{StAR} & 26.4 & 18.7 & 28.5 & 38.9 &  22.4 & 18.9 & 29.8 & 41.6 \\
  \midrule
  KG-BERT~\cite{KG-BERT} & 20.1 & 13.4 & 22.7 & 36.2 & 20.3 & 13.9& 20.1 & 40.3  \\
  KGT5~\cite{KGT5} & 22.5 & 16.6 & 25.9 & 34.6 & 23.4 & 18.1 & 26.2 & 34.3 \\
  \midrule
  BM25+MRC & 23.1 & 17.5 & 25.8 & 34.1 & 25.1 & 20.0 & 27.0 & 36.6 \\
  RAG~\cite{RAG} & 23.6 & 16.4 & 25.0 & 34.3 & 27.8 & 20.6 & 28.6 & 37.7 \\
  IR4KGC & 27.4 & 21.5 & 30.9 & 39.9 & 30.5 & 24.6 & 33.7 & 42.3 \\
  IR4KGC(TuckER) & \textbf{32.1} & \textbf{25.3} & \textbf{35.5} & \textbf{46.2} & \textbf{33.7} & \textbf{25.5} & \textbf{37.7} & \textbf{50.1} \\
  \bottomrule
  \end{tabular}
  }
  \vspace{-0.1cm}
  \caption{Link prediction results on KGC datasets. All metrics are multiplied by 100. The best score is in \textbf{bold}.}
  \label{tab:link_prediction}
  \vspace{-0.2cm}
  \end{table*}

\subsection{Conditional Generation}
\label{sec:conditional_generation}

For every triple query, we use our retrieval module to retrieve the most similar $K$ documents. After that, we use the $\text{F}_L$ function to get the search query and splice it with $K$ documents to form the final input, which is fed into a BART model to generate the predicted entity label. Following~\cite{RAG}, we use language model loss from BART to jointly optimize the retrieval and generation modules. During inference, we use beam search to generate the labels of the predicted entities. Besides, we calculate the edit distance between the predicted label and the label of each entity, and use the entity with the smallest distance as the predicted entity.

\section{Experiments}

\subsection{Experiment Setup}

\noindent{\textbf{Baseline Models}.}  We chose three types of baseline models: (1) KGE models, including RotatE~\cite{RotatE}, TuckER~\cite{TuckER} and StAR~\cite{StAR}. Specifically, StAR uses entity description information; (2) PLM-based KGC model, including KG-BERT~\cite{KG-BERT} and KGT5~\cite{KGT5}; (3) OpenQA models, including BM25+MRC and RAG~\cite{RAG}.
Considering the advantages of the KGE model on inferable relations, we also propose a variant of our model, IR4KGC(TuckER). Specifically, we decide whether to use the results of our model or TuckER on each relation in the test set, based on the performance in the validation set.

\noindent{\textbf{Datasets and Evaluation Metrics}.} In our experiments, we use two datasets based on Wikidata and Freebase, namely CoDEx-M~\cite{codex} and FB15K-237-N~\cite{PKGC}.  To ensure fairness of the comparison, we guarantee that none of the triples in the CoDEx-M test set are present in the pre-training of our retrieval module. Following~\citet{ConvE}, we use MRR and Hits@10 as our evaluation metrics. 

\noindent{\textbf{Implementation Details}.} For our model~\footnote{\url{https://github.com/THU-KEG/IR4KGC}} and BM25+MRC, we use BART-Large as the generative model. In addition, we set the Beam Size to $64$, $N$ to $25$ and $K$ to $5$, respectively.
Please refer to Appendix~\ref{sec:implementation_details} for more implementation details.

\begin{table}[t]
\small
\centering
\setlength{\belowcaptionskip}{-1pt}
\scalebox{0.95}{
    \begin{tabular}{l|cc|cc}
    \toprule
    \multirow{2}{*}{} & \multicolumn{2}{c|}{\textbf{CoDEx-M}} & \multicolumn{2}{c}{\textbf{FB15K-237-N}} \\
    & MRR & Hits@1  & MRR & Hits@1  \\
    \midrule
    TuckER  & 23.7/\textbf{33.8}  & 33.8/\textbf{47.4}  & 14.2/\textbf{38.6} & 25.3/\textbf{53.0}   \\
    KGT5  & 24.1/27.8  & 33.5/40.8  & 16.2/34.2   & 27.8/43.5   \\
    Ours  & \textbf{28.4}/27.3  & \textbf{37.5}/39.3  & \textbf{22.9}/35.3   & \textbf{34.5}/42.9  \\
    \bottomrule
    \end{tabular}
    }
    \vspace{-0.06in}
    \caption{\label{tab:relation_analysis} Link prediction on different types of relations. The front and back of the slash represent the results on uninferable and inferable relations, respectively.}
\end{table}

\subsection{Link Prediction Results}

The link prediction results on two datasets are presented in Table~\ref{tab:link_prediction}. From the table, we can get the following conclusions: (1) KGE models usually outperform the other two types of models. One possible reason is that KGE models are good at handling inferable relations, which account for a large portion of KGs. (2) Comparing the models based on BM25 and Dense Retrieval (RAG), the performance of our model is significantly better. This is because our knowledge retrieval module can retrieve more relevant documents. For detailed examples, please refer to the Appendix~\ref{sec:case_study}. (3) Although our retrieval module is pre-trained on Wikidata, it still achieves good results on the Freebase dataset. This indicates that our retrieval module is generalizable. (4) IR4KGC(TuckER) combines the advantages of TuckER and our model and achieves the best results, which indicates that the future of KGC should be based on multiple models, since they may be good at handling different relations.

\subsection{Relation Analysis}

To further analyze the performance gap of KGC models on different relations, we manually annotate some inferable relations and uninferable relations, and compose the corresponding test sets. Please refer to Appendix~\ref{sec:relation_annotation} for detailed information. The experimental results are shown in Table~\ref{tab:relation_analysis}, which confirm that the KGE model is good at handling inferable relations, while our model is good at uninferable relations. This suggests the need for aggregation between different KGC models.

\section{Conclusion}

In this paper, we find that KGE models struggle with uninferable relations. To solve this problem, we propose a novel KGC model based on information retrieval. Compared with KGE models, the retrieved documents can provide useful information and increase the interpretability of our model. The experimental results show that our model proves to be good at uninferable relations. Besides, the combination of our model and the KGE model achieves the best results. In the future, we will explore the application of our model to temporal KGs.

\section*{Limitation}
This work is based on the English Wikipedia and Wikidata, but can be adapted to most major languages by simply retraining using the corresponding language versions of Wikipedia and Wikidata. In addition, due to the large space occupied by the index, this work has some requirements for the device. It means that our proposed model needs to be deployed on a device with large storage space and memory. Finally, our model also has some requirements for the applied knowledge graph, i.e., the entities and relationships in the KG should have the corresponding label information.

\section*{Ethics Statement}

The data used in this work are drawn from publicly published encyclopedias and knowledge graphs. Most of them are constructed by collaborative editing from a wide range of users and do not involve sensitive data.

\bibliography{anthology,custom}

\begin{thebibliography}{27}
\expandafter\ifx\csname natexlab\endcsname\relax\def\natexlab#1{#1}\fi

\bibitem[{Asai et~al.(2019)Asai, Hashimoto, Hajishirzi, Socher, and
  Xiong}]{asai2019learning}
Akari Asai, Kazuma Hashimoto, Hannaneh Hajishirzi, Richard Socher, and Caiming
  Xiong. 2019.
\newblock Learning to retrieve reasoning paths over wikipedia graph for
  question answering.
\newblock In \emph{International Conference on Learning Representations}.

\bibitem[{Bala{\v{z}}evi{\'c} et~al.(2019)Bala{\v{z}}evi{\'c}, Allen, and
  Hospedales}]{TuckER}
Ivana Bala{\v{z}}evi{\'c}, Carl Allen, and Timothy Hospedales. 2019.
\newblock Tucker: Tensor factorization for knowledge graph completion.
\newblock In \emph{Proceedings of the 2019 Conference on Empirical Methods in
  Natural Language Processing and the 9th International Joint Conference on
  Natural Language Processing (EMNLP-IJCNLP)}, pages 5185--5194.

\bibitem[{Bollacker et~al.(2008)Bollacker, Evans, Paritosh, Sturge, and
  Taylor}]{Freebase}
Kurt Bollacker, Colin Evans, Praveen Paritosh, Tim Sturge, and Jamie Taylor.
  2008.
\newblock Freebase: a collaboratively created graph database for structuring
  human knowledge.
\newblock In \emph{SIGMOD}.

\bibitem[{Bordes et~al.(2013)Bordes, Usunier, Garcia-Duran, Weston, and
  Yakhnenko}]{TransE}
Antoine Bordes, Nicolas Usunier, Alberto Garcia-Duran, Jason Weston, and Oksana
  Yakhnenko. 2013.
\newblock Translating embeddings for modeling multi-relational data.
\newblock In \emph{NeurIPS}.

\bibitem[{Chen et~al.(2017)Chen, Fisch, Weston, and Bordes}]{chen2017reading}
Danqi Chen, Adam Fisch, Jason Weston, and Antoine Bordes. 2017.
\newblock Reading wikipedia to answer open-domain questions.
\newblock In \emph{Proceedings of the 55th Annual Meeting of the Association
  for Computational Linguistics (Volume 1: Long Papers)}, pages 1870--1879.

\bibitem[{Dettmers et~al.(2018)Dettmers, Minervini, Stenetorp, and
  Riedel}]{ConvE}
Tim Dettmers, Pasquale Minervini, Pontus Stenetorp, and Sebastian Riedel. 2018.
\newblock Convolutional 2d knowledge graph embeddings.
\newblock In \emph{Thirty-second AAAI conference on artificial intelligence}.

\bibitem[{Devlin et~al.(2019)Devlin, Chang, Lee, and Toutanova}]{BERT}
Jacob Devlin, Ming-Wei Chang, Kenton Lee, and Kristina Toutanova. 2019.
\newblock Bert: Pre-training of deep bidirectional transformers for language
  understanding.
\newblock In \emph{Proceedings of the 2019 Conference of the North American
  Chapter of the Association for Computational Linguistics}, pages 4171--4186.

\bibitem[{Dong et~al.(2014)Dong, Gabrilovich, Heitz, Horn, Lao, Murphy,
  Strohmann, Sun, and Zhang}]{KGVault}
Xin Dong, Evgeniy Gabrilovich, Geremy Heitz, Wilko Horn, Ni~Lao, Kevin Murphy,
  Thomas Strohmann, Shaohua Sun, and Wei Zhang. 2014.
\newblock Knowledge vault: A web-scale approach to probabilistic knowledge
  fusion.
\newblock In \emph{Proceedings of the 20th ACM SIGKDD international conference
  on Knowledge discovery and data mining}, pages 601--610.

\bibitem[{Guu et~al.(2020)Guu, Lee, Tung, Pasupat, and Chang}]{REALM}
Kelvin Guu, Kenton Lee, Zora Tung, Panupong Pasupat, and Mingwei Chang. 2020.
\newblock Retrieval augmented language model pre-training.
\newblock In \emph{International Conference on Machine Learning}, pages
  3929--3938. PMLR.

\bibitem[{Han et~al.(2018)Han, Cao, Lv, Lin, Liu, Sun, and Li}]{OpenKE}
Xu~Han, Shulin Cao, Xin Lv, Yankai Lin, Zhiyuan Liu, Maosong Sun, and Juanzi
  Li. 2018.
\newblock Openke: An open toolkit for knowledge embedding.
\newblock In \emph{Proceedings of the 2018 conference on empirical methods in
  natural language processing: system demonstrations}, pages 139--144.

\bibitem[{Henderson et~al.(2017)Henderson, Al-Rfou, Strope, Sung, Luk{\'a}cs,
  Guo, Kumar, Miklos, and Kurzweil}]{henderson2017efficient}
Matthew Henderson, Rami Al-Rfou, Brian Strope, Yun-Hsuan Sung, L{\'a}szl{\'o}
  Luk{\'a}cs, Ruiqi Guo, Sanjiv Kumar, Balint Miklos, and Ray Kurzweil. 2017.
\newblock Efficient natural language response suggestion for smart reply.
\newblock \emph{arXiv preprint arXiv:1705.00652}.

\bibitem[{Johnson et~al.(2019)Johnson, Douze, and J{\'e}gou}]{faiss}
Jeff Johnson, Matthijs Douze, and Herv{\'e} J{\'e}gou. 2019.
\newblock Billion-scale similarity search with gpus.
\newblock \emph{IEEE Transactions on Big Data}, 7(3):535--547.

\bibitem[{Karpukhin et~al.(2020)Karpukhin, Oguz, Min, Lewis, Wu, Edunov, Chen,
  and Yih}]{DPR}
Vladimir Karpukhin, Barlas Oguz, Sewon Min, Patrick Lewis, Ledell Wu, Sergey
  Edunov, Danqi Chen, and Wen-tau Yih. 2020.
\newblock Dense passage retrieval for open-domain question answering.
\newblock In \emph{Proceedings of the 2020 Conference on Empirical Methods in
  Natural Language Processing (EMNLP)}, pages 6769--6781.

\bibitem[{Levy et~al.(2017)Levy, Seo, Choi, and Zettlemoyer}]{levy2017zero}
Omer Levy, Minjoon Seo, Eunsol Choi, and Luke Zettlemoyer. 2017.
\newblock Zero-shot relation extraction via reading comprehension.
\newblock In \emph{Proceedings of the 21st Conference on Computational Natural
  Language Learning (CoNLL 2017)}, pages 333--342.

\bibitem[{Lewis et~al.(2020)Lewis, Perez, Piktus, Petroni, Karpukhin, Goyal,
  K{\"u}ttler, Lewis, Yih, Rockt{\"a}schel et~al.}]{RAG}
Patrick Lewis, Ethan Perez, Aleksandra Piktus, Fabio Petroni, Vladimir
  Karpukhin, Naman Goyal, Heinrich K{\"u}ttler, Mike Lewis, Wen-tau Yih, Tim
  Rockt{\"a}schel, et~al. 2020.
\newblock Retrieval-augmented generation for knowledge-intensive nlp tasks.
\newblock \emph{Advances in Neural Information Processing Systems},
  33:9459--9474.

\bibitem[{Lin et~al.(2016)Lin, Liu, and Sun}]{KR-EAR}
Yankai Lin, Zhiyuan Liu, and Maosong Sun. 2016.
\newblock Knowledge representation learning with entities, attributes and
  relations.
\newblock In \emph{IJCAI}, pages 2866--2872.

\bibitem[{Lv et~al.(2022)Lv, Lin, Cao, Hou, Li, Liu, Li, and Zhou}]{PKGC}
Xin Lv, Yankai Lin, Yixin Cao, Lei Hou, Juanzi Li, Zhiyuan Liu, Peng Li, and
  Jie Zhou. 2022.
\newblock Do pre-trained models benefit knowledge graph completion? a reliable
  evaluation and a reasonable approach.
\newblock In \emph{Findings of the Association for Computational Linguistics:
  ACL 2022}, pages 3570--3581.

\bibitem[{Min et~al.(2019)Min, Chen, Zettlemoyer, and
  Hajishirzi}]{min2019knowledge}
Sewon Min, Danqi Chen, Luke Zettlemoyer, and Hannaneh Hajishirzi. 2019.
\newblock Knowledge guided text retrieval and reading for open domain question
  answering.
\newblock \emph{arXiv preprint arXiv:1911.03868}.

\bibitem[{Nickel et~al.(2011)Nickel, Tresp, and Kriegel}]{RESCAL}
Maximilian Nickel, Volker Tresp, and Hans-Peter Kriegel. 2011.
\newblock A three-way model for collective learning on multi-relational data.
\newblock In \emph{Icml}.

\bibitem[{Robertson et~al.(2009)Robertson, Zaragoza et~al.}]{BM25}
Stephen Robertson, Hugo Zaragoza, et~al. 2009.
\newblock The probabilistic relevance framework: Bm25 and beyond.
\newblock \emph{Foundations and Trends{\textregistered} in Information
  Retrieval}, 3(4):333--389.

\bibitem[{Safavi and Koutra(2020)}]{codex}
Tara Safavi and Danai Koutra. 2020.
\newblock Codex: A comprehensive knowledge graph completion benchmark.
\newblock In \emph{Proceedings of the 2020 Conference on Empirical Methods in
  Natural Language Processing (EMNLP)}, pages 8328--8350.

\bibitem[{Saxena et~al.(2022)Saxena, Kochsiek, and Gemulla}]{KGT5}
Apoorv Saxena, Adrian Kochsiek, and Rainer Gemulla. 2022.
\newblock Sequence-to-sequence knowledge graph completion and question
  answering.
\newblock In \emph{Proceedings of the 60th Annual Meeting of the Association
  for Computational Linguistics (Volume 1: Long Papers)}, pages 2814--2828.

\bibitem[{Socher et~al.(2013)Socher, Chen, Manning, and Ng}]{NTN}
Richard Socher, Danqi Chen, Christopher~D Manning, and Andrew Ng. 2013.
\newblock Reasoning with neural tensor networks for knowledge base completion.
\newblock \emph{Advances in neural information processing systems}, 26.

\bibitem[{Sun et~al.(2019)Sun, Deng, Nie, and Tang}]{RotatE}
Zhiqing Sun, Zhi-Hong Deng, Jian-Yun Nie, and Jian Tang. 2019.
\newblock Rotate: Knowledge graph embedding by relational rotation in complex
  space.
\newblock In \emph{ICLR}.

\bibitem[{Wang et~al.(2021)Wang, Shen, Long, Zhou, Wang, and Chang}]{StAR}
Bo~Wang, Tao Shen, Guodong Long, Tianyi Zhou, Ying Wang, and Yi~Chang. 2021.
\newblock Structure-augmented text representation learning for efficient
  knowledge graph completion.
\newblock In \emph{Proceedings of the Web Conference 2021}, pages 1737--1748.

\bibitem[{Wolf et~al.(2020)Wolf, Debut, Sanh, Chaumond, Delangue, Moi, Cistac,
  Rault, Louf, Funtowicz et~al.}]{wolf2020transformers}
Thomas Wolf, Lysandre Debut, Victor Sanh, Julien Chaumond, Clement Delangue,
  Anthony Moi, Pierric Cistac, Tim Rault, R{\'e}mi Louf, Morgan Funtowicz,
  et~al. 2020.
\newblock Transformers: State-of-the-art natural language processing.
\newblock In \emph{Proceedings of the 2020 conference on empirical methods in
  natural language processing: system demonstrations}, pages 38--45.

\bibitem[{Yao et~al.(2019)Yao, Mao, and Luo}]{KG-BERT}
Liang Yao, Chengsheng Mao, and Yuan Luo. 2019.
\newblock Kg-bert: Bert for knowledge graph completion.
\newblock \emph{arXiv preprint arXiv:1909.03193}.

\end{thebibliography}
\bibliographystyle{acl_natbib}

\phantom{a} \\ \phantom{a} \\ \phantom{a} \\ \phantom{a} \\
\phantom{a} \\ \phantom{a} \\ \phantom{a} \\ \phantom{a} \\
\phantom{a} \\ \phantom{a} \\ \phantom{a} \\ \phantom{a} \\
\phantom{a} \\ \phantom{a} \\ \phantom{a} \\ \phantom{a} \\
\phantom{a} \\ \phantom{a} \\ \phantom{a} \\ \phantom{a} \\
\phantom{a} \\ \phantom{a} \\ \phantom{a} \\ \phantom{a} \\
\phantom{a} \\ \phantom{a} \\ \phantom{a} \\ \phantom{a} \\
\phantom{a} \\ \phantom{a} \\ \phantom{a} \\ \phantom{a} \\
\phantom{a} \\ \phantom{a} \\ \phantom{a} \\ \phantom{a} \\
\phantom{a} \\ \phantom{a} \\ \phantom{a} \\ \phantom{a} \\
\phantom{a} \\ \phantom{a} \\ \phantom{a} \\ \phantom{a} \\
\phantom{a} \\ \phantom{a}

\appendix

\begin{table*}[hbt!]
  \centering
  \hspace*{-0.32cm}
  \setlength\tabcolsep{5.5pt}
  \scalebox{0.8}{
  \begin{tabular}{lcc}
  \toprule
    Dataset & Inferable Relations &  Uninferable Relations \\ 
  \midrule
  \multirow{7}{*}{CoDEx-M} & \textit{country} (P17) & \textit{cause of death} (P509)  \\
  & \textit{languages spoken, written or signed} (P1412) & \textit{place of burial} (P119)   \\
  & \textit{occupation} (P106) & \textit{place of birth} (P19)  \\
  & \textit{country of citizenship} (P27) & \textit{place of death} (P20)  \\
  & \textit{member of} (P463) & \textit{religion or worldview} (P140)  \\
  & \textit{located in the administrative territorial entity} (P131) & \textit{child} (P40)  \\
  & \textit{headquarters location} (P159) & \textit{director} (P57)  \\
  \midrule
  \multirow{7}{*}{FB15K-237-N}   & \textit{/influence/influence\_node/influenced\_by} & \textit{/people/cause\_of\_death/people}  \\ 
  & \textit{/location/location/time\_zones} & \textit{/people/person/place\_of\_birth}  \\
  & \textit{/people/person/nationality} & \textit{/film/film/featured\_film\_locations}  \\
  & \textit{/location/country/official\_language} & \textit{/people/deceased\_person/place\_of\_death}  \\
  & \textit{/people/ethnicity/languages\_spoken} & \textit{/music/artist/origin}  \\
  & \textit{/sports/sports\_team/sport} & \textit{/education/educational\_institution/colors}  \\
  & \textit{/tv/tv\_program/genre} & \textit{/organization/organization\_founder/organizations\_founded}  \\
  \bottomrule
  \end{tabular}
  }
  \vspace{-0.1cm}
  \caption{Inferable relations and uninferable relations that we annotate on the KGC datasets.}
  \label{tab:relation_annotation}
  \vspace{-0.2cm}
\end{table*}

\section{Implementation Details}
\label{sec:implementation_details}

For Wikipedia and Wikidata, we use the snapshots of 20201009 and 20210414, respectively, to construct our pre-trained corpus. We pre-train our retrieval module based on the codes~\footnote{\url{https://github.com/facebookresearch/DPR}} of~\citet{DPR}. We use its default parameters and set the maximum number of training epochs to 3. For RotatE, we use the code provided by~\citet{OpenKE}. For RAG, we use the code provided by~\citet{wolf2020transformers}. For the other models, we use the default code from the paper. For the implementation of BM25+MRC, we use BM25 to retrieve relevant documents, which are concatenated with the search query as input to BART to generate predicted answers. For RAG and BM25+MRC, we use the same method mentioned in Section~\ref{sec:conditional_generation} to map the predicted labels to the corresponding entities. For our model, we train up to 20 epochs. For other models, we use their default maximum number of training epochs. In addition, we use Hits@10 to select the best model on the validation set.

\section{Case Study}
\label{sec:case_study}

In Table~\ref{tab:case_study}, we provide the most relevant documents retrieved by our model, RAG and BM25+MRC for different triple queries. From the table, we can see that our retrieval module has the best retrieval results among the three models. It is also worth noting that the Dense Retrieval used by RAG is failed after fine-tuning on the KGC task.

\section{Relation Annotation}
\label{sec:relation_annotation}

Whether a relation is inferable or not is a subjective question. Therefore, we invite some graduate students to label their own perceived inferable relations and uninferable relation on the KGC datasets we use. In the end, the seven relations with the most votes are retained separately. The information about the relations we finally selected is listed in Table~\ref{tab:relation_annotation}.

\section{Computing Infrastructure}

We use a machine with 512G RAM, 72 core CPUs and 8 Tesla V100s to deploy our models and run experiments.

 \begin{table*}[t]
  \centering
  \hspace*{-0.32cm}
\setlength\tabcolsep{5.5pt}
  \scalebox{0.9}{
  \begin{tabular}{ccp{10cm}}
    \toprule
    Triple Query & Model &  \makebox[10cm][c]{Retrieved Document}  \\ 
  \midrule
  \multirow{17}{*}{(Lebron James, \textit{place of birth}, ?)} & \multirow{2}{*}{BM25+MRC}  &  This page lists career accomplishments of the American professional basketball player LeBron James. \\
  \cmidrule(lr){2-3}
  & \multirow{8}{*}{RAG} & Fiasco was born Wasalu Muhammad Jaco on February 16, 1982 in Chicago, Illinois. Fiasco is of West African descent, one of nine children of Shirley, a gourmet chef, and Gregory, an engineer. His father, a member of the Black Panther Party, was a prolific African drummer, karate teacher, operating plant engineer, and owner of karate schools and army surplus stores. Fiasco was raised Muslim on the West Side of Chicago on Madison Terrace housing project. At the age of three, Fiasco began taking martial arts classes.   \\
  \cmidrule(lr){2-3}
  & \multirow{7}{*}{IR4KGC} &  James was selected to the All-NBA First Team twelve times (all-time record), made the All-Defensive First Team five times, and has played in sixteen All-Star Games, in which he was selected All-Star MVP three times. \textbf{James played basketball for St. Vincent–St. Mary High School in his hometown of Akron, Ohio}, where he was heavily touted by the national media as a future NBA superstar. A prep-to-pro, he joined Cleveland in 2003 as the first draft pick.  \\
  \midrule
  \multirow{27}{*}{(Avatar, \textit{director}, ?)} & \multirow{8}{*}{BM25+MRC}  &  On April 27, 2017, the release dates for all four sequels were ultimately announced: December 18, 2020 for Avatar 2, December 17, 2021 for Avatar 3, December 20, 2024 for Avatar 4, and December 19, 2025 for Avatar 5. The films will be released in Dolby Vision. \textbf{New crew members include cinematographer Russell Carpenter, who worked with Cameron on True Lies and Titanic}, and Aashrita Kamath, who will act as art director on all four sequels. \\
  \cmidrule(lr){2-3}
  & \multirow{10}{*}{RAG} & UiPath was founded in 2005 by the Romanian entrepreneurs, Daniel Dines and Marius Tirca, is a global software company that develops a platform for Robotic Process Automation (RPA or RPAAI). The company started from Bucharest, Romania and later opened offices in London, New York, Bengaluru, Singapore, and Tokyo. In 2017, the company reported 590 employees and moved its headquarters to New York to be closer to its international customer base, which rose to 700 customers from 100 in 2016. In August 2015, UiPath closed an initial seed funding round of \$1.6 million led by the Earlybird Venture Capital.   \\
  \cmidrule(lr){2-3}
  & \multirow{9}{*}{IR4KGC} &  \textbf{Avatar (marketed as James Cameron's Avatar)} is a 2009 American epic science fiction \textbf{film directed, written, produced, and co-edited by James Cameron} and stars Sam Worthington, Zoe Saldana, Stephen Lang, Michelle Rodriguez, and Sigourney Weaver. The film is set in the mid-22nd century when humans are colonizing Pandora, a lush habitable moon of a gas giant in the Alpha Centauri star system, in order to mine the mineral unobtanium, a room-temperature superconductor.This property of Unobtanium is stated in movie guides, rather than in the film.  \\
  \bottomrule
  \end{tabular}
  }
  \vspace{-0.1cm}
  \caption{For different triple queries, we give the most relevant document retrieved by different models. For display purposes, we have \textbf{bolded} the parts of the document that contain the answer to the corresponding triple query.}
  \label{tab:case_study}
  \vspace{-0.2cm}
  \end{table*}

\end{document}